\documentclass{llncs}
\usepackage{lmodern}

\usepackage[utf8]{inputenc}
\usepackage{bm}
\usepackage{amsmath}
\usepackage{amssymb}
\usepackage{graphicx}
\usepackage{hyperref}
\usepackage{esvect}
\usepackage{listings}
\lstdefinelanguage{json}{
  basicstyle=\ttfamily,
  numbers=left,
  numberstyle=\tiny, 
  stepnumber=1,
  numbersep=5pt,
  showstringspaces=false,
  breaklines=true,
  frame=lines,
  backgroundcolor=\color{white},
  literate={"}{{\texttt{\char34}}}1
           {,}{{\texttt{,}}}1
           {:}{{\texttt{:}}}1
           {[}{{\texttt{[}}}1
           {]}{{\texttt{]}}}1
           {\{}{{\texttt{\{}}}1
           {\}}{{\texttt{\}}}}1
}
\usepackage{xcolor}
\usepackage{booktabs}

\title{PYTHEN: A Flexible Framework for Legal Reasoning in Python}

\author{Nguyen Ha Thanh\inst{1,2} \and Ken Satoh\inst{1}}

\institute{
Center for Juris-Informatics, ROIS-DS, Tokyo, Japan\and
Research and Development Center for Large Language Models, NII, Tokyo, Japan
}

\begin{document}

\maketitle

\begin{abstract}
This paper introduces PYTHEN, a novel Python-based framework for defeasible legal reasoning. PYTHEN is designed to model the inherently defeasible nature of legal argumentation, providing a flexible and intuitive syntax for representing legal rules, conditions, and exceptions. Inspired by PROLEG (PROlog-based LEGal reasoning support system) and guided by the philosophy of The Zen of Python, PYTHEN leverages Python's built-in \texttt{any()} and \texttt{all()} functions to offer enhanced flexibility by natively supporting both conjunctive (ALL) and disjunctive (ANY) conditions within a single rule, as well as a more expressive exception-handling mechanism. This paper details the architecture of PYTHEN, provides a comparative analysis with PROLEG, and discusses its potential applications in autoformalization and the development of next-generation legal AI systems. By bridging the gap between symbolic reasoning and the accessibility of Python, PYTHEN aims to democratize formal legal reasoning for young researchers, legal tech developers, and professionals without extensive logic programming expertise. We position PYTHEN as a practical bridge between the powerful symbolic reasoning capabilities of logic programming and the rich, ubiquitous ecosystem of Python, making formal legal reasoning accessible to a broader range of developers and legal professionals.

\end{abstract}

\section{Introduction}

The field of AI and Law has long pursued the goal of creating computational models of legal reasoning. A central challenge in this endeavor is capturing the nuanced and defeasible nature of legal argumentation, where rules are seldom absolute and are frequently subject to exceptions and qualifications. Traditional programming paradigms often fall short in representing this complexity in a clear and maintainable way.

Logic programming, with Prolog as its most prominent example, has offered a powerful paradigm for this task. Systems like PROLEG (PROlog-based LEGal reasoning support system) have demonstrated the viability of using logic programming to model sophisticated legal theories, such as the Japanese Presupposed Ultimate Fact (JUF) Theory. These systems excel at handling the burden of proof and reasoning with incomplete information, which are hallmarks of legal practice.

However, the adoption of Prolog-based systems has been hampered by several significant barriers. First, the steep learning curve of Prolog makes it inaccessible to many researchers and developers, particularly young researchers entering the field of legal AI who may lack formal training in logic programming. Second, the difficulty of integrating Prolog systems into the broader ecosystem of modern software development, which is largely dominated by languages like Python, creates a practical gap between the powerful theoretical models developed in the AI and Law community and the tools available to legal tech developers and practitioners. This creates a significant accessibility barrier that limits the adoption and impact of formal legal reasoning systems.

To address this gap, we propose \textbf{PYTHEN}\footnote{\url{https://github.com/nguyenthanhasia/pythen}}, a Python-native framework for defeasible legal reasoning. PYTHEN is designed to be both powerful and easy to use, providing a clear and expressive syntax for defining legal rules and their exceptions. It allows for complex conditions (both \texttt{ALL} and \texttt{ANY}) within a single rule structure, a feature that enhances its flexibility compared to more rigid logic programming formalisms.

The design philosophy of PYTHEN is deeply influenced by The Zen of Python (PEP 20), which emphasizes principles such as ``Beautiful is better than ugly,'' ``Simple is better than complex,'' and ``Readability counts.'' By adhering to these principles, PYTHEN ensures that legal rules are not only formally correct but also human-readable and maintainable, even for those without extensive training in formal logic.

This paper makes the following contributions:

\begin{enumerate}
\item It introduces the PYTHEN framework, detailing its rule and fact structure, and explaining how it leverages Python's built-in \texttt{any()} and \texttt{all()} functions.
\item It provides a comparative analysis of PYTHEN and PROLEG, highlighting the advantages of PYTHEN's flexible syntax and accessibility.
\item It discusses the potential of PYTHEN in the context of autoformalization and its integration with large language models (LLMs), particularly for converting natural language legal texts to structured rules.
\item It situates PYTHEN within the broader landscape of legal AI research, drawing connections to work from leading conferences and journals like ICAIL, JURIX, and the AI and Law journal, and discussing recent efforts to bridge NLP and formal legal reasoning.
\item It emphasizes the accessibility and democratization aspects of PYTHEN, particularly for young researchers and professionals without extensive logic programming expertise.
\end{enumerate}

By leveraging the ubiquity and flexibility of Python, combined with its philosophical emphasis on simplicity and readability, PYTHEN aims to democratize the development of sophisticated legal reasoning systems, enabling a new generation of legal tech applications and making formal legal reasoning accessible to a broader audience.

\section{Related Work}

The development of PYTHEN builds upon a rich history of research in AI and Law, particularly in the areas of logic-based legal reasoning, defeasible argumentation, natural language processing for legal texts, and autoformalization. This section reviews the key influences and situates our work within the current state of the art.

\subsection{Logic-Based Legal Reasoning and PROLEG}

The idea of representing legal statutes as a set of logical rules dates back to the early days of AI and Law. One of the most influential systems in this tradition is \textbf{PROLEG} (PROlog-based LEGal reasoning support system), developed by Ken Satoh and his colleagues \cite{satoh2010proleg}. PROLEG is an implementation of the Japanese ``theory of presupposed ultimate facts'' (JUF theory), a sophisticated model of judicial reasoning that deals with the allocation of the burden of proof under conditions of incomplete information. The system uses a Prolog-based engine to determine whether a legal claim is justified given a set of facts and rules. PROLEG's strength lies in its ability to model the defeasible nature of legal rules, where a conclusion can be overturned by exceptions.

Extensions to PROLEG, such as \textbf{ArgPROLEG} \cite{shams2013argproleg}, have sought to integrate it with formal argumentation frameworks, making the reasoning process more transparent and explainable. These works have demonstrated the power of logic programming for modeling complex legal doctrines. However, their reliance on Prolog has limited their accessibility to the broader legal tech community and to young researchers without formal training in logic programming.

\subsection{Defeasible Logic and Argumentation in Law}

Legal reasoning is inherently defeasible, a point that has been extensively argued in the literature \cite{prakken2015law,walton2005argumentation}. A legal conclusion is always provisional, subject to defeat by new evidence or arguments. This has led to the development of various formalisms for defeasible reasoning, such as Defeasible Logic \cite{nute2001defeasible} and formal argumentation theory \cite{dung1995acceptability}. These approaches provide a formal basis for modeling legal disputes as a dialogue between competing arguments, where the ultimate outcome depends on which arguments can withstand attack.

The International Conference on Artificial Intelligence and Law (ICAIL) and the JURIX conferences have been major venues for the presentation of research in this area. The work presented at these conferences has shown the importance of modeling not just the rules themselves, but also the process of argumentation and the allocation of the burden of proof \cite{prakken2007formalising}. PYTHEN draws on these insights by providing a clear and explicit mechanism for representing both conditions and exceptions within its rule structure.

\subsection{Legal Ontologies and Knowledge Representation}

The formalization of legal knowledge through ontologies has been a significant research direction. The LKIF Core Ontology \cite{hoekstra2007lkif} provides a foundational framework for representing basic legal concepts in a machine-readable format. Subsequent work on legal ontologies \cite{valente2005types,valente1994functional} has established methodologies for capturing legal domain knowledge in ways that support reasoning and knowledge reuse.

More recently, legal knowledge graphs have emerged as a powerful approach to representing and reasoning over legal information. Projects such as LYNX \cite{rodriguez2020lynx} demonstrate how semantic web technologies can be applied to legal data across multiple jurisdictions and languages. These approaches complement PYTHEN by providing structured representations of legal concepts that can be integrated with PYTHEN's rule-based reasoning engine.

\subsection{Natural Language Processing and Autoformalization of Legal Text}

A significant bottleneck in the development of rule-based legal AI systems has been the manual process of formalizing legal texts into a machine-readable format. \textbf{Autoformalization}---the automatic translation of natural language into a formal representation---has emerged as a promising solution to this problem \cite{wu2022autoformalization,mensfelt2025towards}. The advent of large language models (LLMs) has brought new momentum to this area, with recent work demonstrating the potential of LLMs to translate mathematical and legal texts into formal specifications \cite{ariai2025natural}.

Recent efforts have specifically targeted the conversion of natural language legal texts into PROLEG format. The KRAG Framework \cite{thanh2024krag} introduces Soft PROLEG, an extension of PROLEG designed to work with LLMs, using inference graphs to aid in structured legal reasoning. This work demonstrates that LLMs can be effectively guided to produce PROLEG-compatible outputs when provided with clear structural templates and examples.

Work on improving the translation of case descriptions into logical formulas \cite{zin2024addressing} has shown that domain-specific named entity recognition (NER) techniques can significantly improve the quality of autoformalization. These techniques extract key legal entities and relationships from natural language text, which can then be mapped to formal rule structures.

PYTHEN's JSON-based syntax is particularly well-suited for serving as an intermediate representation in autoformalization pipelines. By providing a clear and well-defined target formalism, PYTHEN can serve as the ``semantic backbone'' for an autoformalization pipeline, where an LLM first generates a draft formalization of a legal text, which is then refined and validated against the PYTHEN rule structure. This hybrid approach, combining the generative power of LLMs with the rigor of symbolic reasoning, represents a promising direction for the future of legal AI \cite{fuchs2024intermediate}.

\subsection{Legal AI Benchmarks and Evaluation}

The emergence of comprehensive benchmarks for legal AI has been crucial for advancing the field. LegalBench \cite{guha2023legalbench} provides a benchmark dataset for evaluating legal task specification and reasoning across multiple dimensions. The Cambridge Law Corpus \cite{ostling2023cambridge} offers a large-scale dataset for legal AI research, with annotations and benchmarks for case outcome prediction and other legal reasoning tasks.

These benchmarks highlight the challenges that legal AI systems must address, including the need for precise knowledge representation, handling of complex legal concepts, and the ability to reason with incomplete or ambiguous information. PYTHEN's design is informed by these challenges and aims to provide a framework that can support the development of systems capable of handling such complexity.

\subsection{Hybrid Approaches: Combining Symbolic and Neural Methods}

Recent research has explored the integration of rule-based symbolic reasoning with neural network approaches \cite{sadowski2025verifiable}. These hybrid systems leverage the interpretability and logical soundness of symbolic methods with the pattern recognition capabilities of neural networks. PYTHEN is positioned as a tool that can facilitate such hybrid approaches, providing a clear interface between symbolic legal reasoning and neural methods like LLMs.

The integration of PYTHEN with LLMs through autoformalization pipelines represents a modern instantiation of this hybrid approach, where LLMs generate candidate rule structures that are then validated and refined using PYTHEN's formal reasoning engine.

\section{The PYTHEN Framework}

PYTHEN is designed with the primary goal of making the formalization of legal rules both intuitive for humans and computationally tractable. It achieves this through a simple, JSON-based structure for representing rules and facts, which can be easily created, parsed, and manipulated using standard Python libraries. The design is guided by the principles of The Zen of Python, particularly the emphasis on simplicity, readability, and accessibility.

\subsection{Design Philosophy: The Zen of Python}

PYTHEN's design is deeply influenced by The Zen of Python (PEP 20), a set of guiding principles for Python development. Key principles that guide PYTHEN's design include:

\begin{itemize}
\item \textbf{Beautiful is better than ugly}: PYTHEN uses a clean, JSON-based syntax that is visually clear and easy to understand.
\item \textbf{Simple is better than complex}: Rather than requiring users to learn Prolog syntax, PYTHEN uses familiar Python concepts.
\item \textbf{Readability counts}: Legal rules in PYTHEN are self-documenting and easy to understand, even for those without formal logic training.
\item \textbf{Explicit is better than implicit}: PYTHEN makes conditions, exceptions, and reasoning steps explicit and visible.
\item \textbf{Practicality beats purity}: PYTHEN prioritizes usability and integration with modern tools over strict formal purity.
\end{itemize}

These principles ensure that PYTHEN is not only formally sound but also accessible and practical for real-world legal tech applications.

\subsection{Rule Structure}

A PYTHEN rule is a dictionary that defines a legal proposition and the conditions under which it holds true. Each rule consists of the following key-value pairs:

\begin{itemize}
\item \texttt{``p''}: The \textbf{proposition} that the rule defines. This is a unique string identifier for the legal concept being modeled (e.g., \texttt{``art17\_erasure\_applicable''}).
\item \texttt{``op''}: The \textbf{operator} that governs the relationship between the conditions. It can be either \texttt{``ALL''} (conjunctive), meaning all conditions must be met, or \texttt{``ANY''} (disjunctive), meaning at least one condition must be met. This design is inspired by Python's built-in \texttt{all()} and \texttt{any()} functions, which are familiar to any Python developer.
\item \texttt{``conditions''}: A list of strings, where each string is a proposition that must be evaluated to determine the truth of the main proposition \texttt{p}. These can be references to other rules or to basic facts.
\item \texttt{``exceptions''}: A list of strings representing propositions that, if true, will defeat the main proposition, even if its conditions are met. This provides a direct mechanism for modeling defeasibility.
\end{itemize}

The use of \texttt{``ALL''} and \texttt{``ANY''} operators directly mirrors Python's built-in functions, making the semantics immediately clear to Python developers. This design choice significantly enhances accessibility for developers who may not be familiar with formal logic notation.

Below is an example of a rule from a PYTHEN rule-base, modeling a part of Article 17 of the GDPR (the ``right to be forgotten''):

\begin{lstlisting}
{
    "p": "art17_erasure_applicable",
    "op": "ANY",
    "conditions": [
        "no_longer_necessary",
        "consent_withdrawn",
        "object_to_processing",
        "processing_unlawful",
        "child_data_collected"
    ],
    "exceptions": [
        "freedom_of_expression",
        "legal_obligation",
        "public_interest_archiving_research",
        "legal_claims"
    ]
}
\end{lstlisting}

In this example, the proposition \texttt{art17\_erasure\_applicable} is true if \texttt{ANY} of its conditions are met (e.g., if consent is withdrawn), provided that none of the \texttt{exceptions} (e.g., the data is needed for a legal claim) are true.

\subsection{Fact Base}

The fact base is a simple list of strings, where each string represents a basic proposition that is considered to be true for a given case. These are the foundational inputs to the reasoning process.

Example Fact Base:

\begin{lstlisting}
[
    "objection_to_direct_marketing",
    "data_collected_from_child",
    "consent_was_basis",
    "consent_is_withdrawn",
    "data_not_needed_for_purpose"
]
\end{lstlisting}

\subsection{Reasoning Mechanism}

The PYTHEN reasoner is a goal-driven engine that works backward from a target proposition. To determine whether a proposition \texttt{p} holds, the engine applies an evaluation strategy over rules, conditions, and exceptions. In the current implementation, the following steps are performed:

\begin{enumerate}
\item \textbf{Exception Evaluation}: The engine first checks whether any proposition listed in the \texttt{exceptions} of \texttt{p} can be derived. If at least one exception is found to hold, the reasoning process for \texttt{p} terminates and \texttt{p} is rejected. This exception-first evaluation constitutes a decision-oriented strategy that allows the reasoner to efficiently compute a final judgement in defeasible settings.

\item \textbf{Condition Evaluation}: If no exception is derived, the engine evaluates the \texttt{conditions} according to the specified operator \texttt{op}.
\begin{itemize}
\item If \texttt{op} is \texttt{``ALL''}, the engine recursively attempts to establish that all propositions in the \texttt{conditions} list hold (corresponding to Python's \texttt{all()} semantics).
\item If \texttt{op} is \texttt{``ANY''}, the engine recursively attempts to establish that at least one proposition in the \texttt{conditions} list holds (corresponding to Python's \texttt{any()} semantics).
\end{itemize}

\item \textbf{Base Case}: The recursion terminates when a proposition is found in the fact base, in which case it is considered true, or when no rule is available to derive it, in which case it is considered false.
\end{enumerate}

It is important to note that the choice of evaluation order affects the structure of the reasoning process and its explanatory trace, but not the logical outcome of the derivation under standard assumptions of defeasible reasoning. While legal practitioners often reason by first establishing the applicability of general rules and subsequently considering exceptions, the exception-first strategy adopted here prioritizes efficient judgement computation. Alternative evaluation orders, including general-rule-first and explanation-oriented strategies, can be supported without changing the underlying rule representation.

\subsection{Evaluation Strategy and Computational Considerations}

An important design choice in defeasible legal reasoning systems concerns the evaluation strategy used to determine whether a proposition holds. While different evaluation orders—such as general-rule-first or exception-first—are often logically equivalent in terms of their final outcomes, they may differ significantly in their computational behavior and explanatory characteristics.

In practice, the computational cost of reasoning over general rules and over exceptions may vary substantially depending on the structure of the rule base, the depth of rule dependencies, and the availability of factual information. General-rule reasoning may involve the evaluation of multiple conjunctive or disjunctive conditions, while exception reasoning may terminate early if a single defeating condition can be established. As a result, a fixed evaluation order may be suboptimal in terms of efficiency across different legal domains and cases.

From a system-oriented perspective, evaluation strategy can therefore be viewed as an execution policy rather than a property of the rule representation itself. PYTHEN explicitly separates the representation of legal rules from their evaluation order, allowing alternative strategies to be explored without modifying the underlying rule base. This design enables the incorporation of computational considerations, such as the relative difficulty of general-rule reasoning versus exceptional reasoning, into the reasoning process.

More adaptive strategies are also possible. For example, general-rule reasoning and exception reasoning can be executed concurrently or in parallel, allowing the reasoning process to terminate as soon as a decisive result is obtained. In such settings, dynamic allocation of computational resources between competing reasoning paths can serve as a practical approximation of computational hardness prediction.

From a jurisprudential perspective, the distinction between general rules and exceptions is not absolute, but rather a methodological device for organizing issues and structuring fact-finding. Legal practice suggests that the preferred order of examination depends on context and legal domain. In civil law, anticipatory judgments based on defenses may be theoretically possible, but they are rarely adopted in practice due to the close interdependence between facts underlying claims and defenses. In criminal law, by contrast, the order of judgment may carry normative significance, making certain anticipatory evaluations inappropriate, as determinations regarding unlawfulness or culpability have independent legal meaning. These observations further support the view that evaluation order should be treated as a context-sensitive strategy rather than a fixed semantic property of legal rules.

\section{Comparison with PROLEG}

While PYTHEN is inspired by the pioneering work of PROLEG, it introduces several key differences in its design and philosophy, aimed at increasing flexibility, accessibility, and practical usability. This section provides a comparative analysis of the two frameworks.

\subsection{Syntax and Expressiveness}

PROLEG, being based on Prolog, inherits its syntax, which is powerful but can be opaque to those not trained in logic programming. A PROLEG rule is typically expressed as a Prolog clause, which, while formally precise, does not always map intuitively to the way a lawyer might articulate a rule. This creates a significant barrier to adoption, particularly for young researchers and legal professionals without formal training in logic.

PYTHEN, in contrast, uses a simple, human-readable JSON structure. This has multiple significant advantages:

\begin{enumerate}
\item \textbf{Accessibility}: The JSON format is universally understood and can be easily generated and parsed by virtually any programming language. This lowers the barrier to entry for developers and legal professionals who may not be familiar with Prolog or formal logic programming. This is particularly important for young researchers entering the field of legal AI.

\item \textbf{Flexibility in Conditions}: A key innovation in PYTHEN is the explicit \texttt{``op''} field, which allows a rule's conditions to be either conjunctive (\texttt{``ALL''}) or disjunctive (\texttt{``ANY''}). In PROLEG, achieving disjunctive conditions typically requires writing multiple, separate rules for the same proposition. PYTHEN's approach allows for a more compact and, in many cases, more natural representation of legal rules that involve alternative paths to a conclusion.

\item \textbf{Integration with Python Ecosystem}: By using Python's \texttt{any()} and \texttt{all()} semantics, PYTHEN creates an intuitive bridge between formal logic and Python programming, making it immediately familiar to the millions of Python developers worldwide.
\end{enumerate}

\subsection{Exception Handling}

Both PROLEG and PYTHEN are designed to model the defeasible nature of legal rules, but they adopt different design perspectives regarding the representation and evaluation of exceptions. PROLEG handles exceptions through negation as failure and rule ordering within a Prolog-based evaluation framework. This approach is closely tied to legal-theoretical considerations and provides well-defined semantics, but it may result in interactions between rules that are not immediately visible at the level of individual rule definitions.

PYTHEN, in contrast, adopts an explicit representation of defeasibility by associating each rule with a dedicated \texttt{``exceptions''} list. This design emphasizes transparency at the knowledge-representation level: the conditions under which a proposition can be defeated are localized within the rule itself, making defeasible relationships directly observable. Importantly, PYTHEN treats the representation of exceptions as orthogonal to the choice of evaluation strategy. Different execution orders or reasoning strategies can be applied without modifying the rule base, allowing the same representation to support both decision-oriented and explanation-oriented reasoning.

\subsection{Ecosystem and Integration}

Perhaps the most significant practical difference lies in the ecosystems of the two frameworks. PROLEG is embedded in the Prolog ecosystem, which, while powerful for symbolic reasoning, is relatively isolated from the mainstream of modern software development.

PYTHEN, being a native Python framework, has immediate access to the vast and rich Python ecosystem. This includes:

\begin{itemize}
\item \textbf{Data Science and NLP Libraries}: Seamless integration with libraries like Pandas, NLTK, and spaCy for pre-processing legal documents and extracting facts. This facilitates the development of end-to-end legal AI systems that combine NLP with formal reasoning.

\item \textbf{Machine Learning Frameworks}: The ability to connect with PyTorch and TensorFlow for building hybrid models that combine symbolic reasoning with machine learning, enabling the development of more sophisticated legal AI systems.

\item \textbf{Web Frameworks}: Easy integration with Django and Flask for building interactive legal tech applications and APIs, making it practical to deploy legal reasoning systems in real-world applications.

\item \textbf{LLM Integration}: Direct access to libraries for interacting with large language models like GPT-4 and Claude, facilitating the development of autoformalization pipelines that convert natural language legal texts into PYTHEN rules.
\end{itemize}

This ease of integration makes PYTHEN a more practical choice for building end-to-end legal AI systems in a modern technology stack, and significantly lowers the barrier to adoption for young researchers and legal tech developers.

\subsection{Accessibility and Democratization}

A critical advantage of PYTHEN over PROLEG is its accessibility to researchers and developers without extensive formal logic training. The legal AI field is rapidly growing, and many young researchers are entering from backgrounds in NLP, machine learning, or software engineering rather than formal logic or Prolog. PYTHEN's design explicitly addresses this gap by using familiar Python concepts and JSON syntax.

Furthermore, PYTHEN's integration with LLMs through autoformalization pipelines makes it possible for researchers to work with formal legal reasoning without manually writing Prolog code. This democratization of formal legal reasoning is a significant contribution to making the field more accessible and inclusive.

\subsection{A Comparative Example}

Consider a simplified rule: ``A contract is voidable if the person was a minor OR was mentally incapable, UNLESS the contract was for necessities.''

In \textbf{PROLEG}, this might be represented by separate rules:

\begin{lstlisting}[language=prolog]
contract_voidable(C) :- minor(P), party_to(P, C).
contract_voidable(C) :- incapable(P), party_to(P, C).
exception(contract_voidable(C),for_necessities(C)).
\end{lstlisting}

In \textbf{PYTHEN}, this can be represented in a single, more intuitive rule:

\begin{lstlisting}
{
    "p": "contract_voidable",
    "op": "ANY",
    "conditions": ["minor", "incapable"],
    "exceptions": ["for_necessities"]
}
\end{lstlisting}

This simple example illustrates how PYTHEN's syntax can lead to a more compact and readable formalization of legal rules, particularly those involving disjunctive conditions. Moreover, a developer without Prolog experience can immediately understand the PYTHEN rule, while the Prolog version requires knowledge of Prolog syntax and semantics.

\section{Applications and Use Cases}

The flexibility and accessibility of PYTHEN open up a wide range of potential applications in legal tech and computational law, particularly for young researchers and developers entering the field.

\subsection{Autoformalization and LLM-Powered Systems}

PYTHEN is well-suited to serve as the target formalism for autoformalization pipelines. An LLM can be prompted to translate a piece of legal text (e.g., a clause from a contract or a section of a statute) into a PYTHEN rule structure. The structured nature of PYTHEN makes it easier to validate and debug the output of the LLM, and the resulting rule base can be executed with the PYTHEN reasoner to ensure logical consistency. This creates a powerful synergy between the generative capabilities of LLMs and the formal rigor of symbolic reasoning.

Young researchers can leverage pre-trained LLMs to rapidly prototype legal reasoning systems without needing to manually write formal rules, significantly lowering the barrier to entry for the field.

\subsection{Compliance and Regulatory Analysis}

Companies in highly regulated industries, such as finance and healthcare, can use PYTHEN to model complex regulatory requirements. For example, a rule base could be created to determine whether a particular financial transaction complies with anti-money laundering (AML) regulations, or whether a proposed use of patient data complies with HIPAA. The clear, JSON-based rules would be auditable by compliance officers, and the reasoning engine could be integrated into business workflows to provide real-time compliance checks.

\subsection{Legal Education and Training}

PYTHEN can be a valuable tool for legal education and training. Law students and young researchers could learn about the logical structure of legal rules by creating their own PYTHEN rule bases for specific areas of law. This would provide them with a hands-on understanding of concepts like conditions, exceptions, and the burden of proof. The interactive nature of the framework would allow them to test different factual scenarios and see how they affect the legal outcome. The simplicity of PYTHEN's syntax makes it accessible to students without extensive programming background.

\subsection{Contract Analysis and Management}

By formalizing the clauses of a contract into a PYTHEN rule base, it becomes possible to automatically analyze the contract for potential issues, such as conflicting clauses or missing provisions. For example, a rule base could be created to check if a software license agreement contains adequate data protection clauses. This could be integrated into contract lifecycle management (CLM) systems to provide automated risk assessment.

\section{Discussion and Future Work}

PYTHEN shows that defeasible legal reasoning can be operationalized in a way that is both practical and compatible with contemporary AI systems. By embedding formal reasoning directly into Python, PYTHEN aligns legal reasoning with data-driven and model-centric workflows, including large language models.

Future work will focus on scaling and orchestration. First, we will study resource allocation strategies for legal reasoning pipelines, including when and how to invoke symbolic reasoning versus statistical or neural components. Second, we will investigate divide-and-conquer approaches that decompose complex legal problems into smaller, independently solvable subproblems, improving efficiency and modularity. Third, we will explore agentic AI settings in which multiple specialized agents—such as text interpretation, rule induction, and defeasible reasoning agents—coordinate using PYTHEN as a shared reasoning substrate. These directions aim to position PYTHEN as a core component in hybrid, multi-agent legal AI systems.

\section{Conclusion}

This paper introduced PYTHEN, a Python-based framework for defeasible legal reasoning that emphasizes simplicity, extensibility, and practical integration. By lowering the barrier to formal legal reasoning and enabling tight coupling with modern AI workflows, PYTHEN supports the development of more transparent, scalable, and reliable legal AI systems. From a theoretical perspective, PYTHEN also provides a concrete platform for studying meta-reasoning issues such as conflict management, theory revision, and control of reasoning resources in hybrid and agent-based legal AI architectures.

\bibliographystyle{splncs04}
\bibliography{references.bib}

\end{document}